# GeoDecoder: Empowering Multimodal Map Understanding

Feng Qi, Mian Dai, Zixian Zheng, Chao Wang

GaoDe Map, Alibaba Group

{qianlong.qf, zhengzixian.zzx}@alibaba-inc.com; { daimian.dm, charles.wang}@autonavi.com

## Abstract

*This paper presents GeoDecoder, a dedicated multimodal model designed for processing geospatial information in maps. Built on the BeitGPT architecture, GeoDecoder incorporates specialized expert modules for image and text processing. On the image side, GeoDecoder utilizes GaoDe Amap as the underlying base map, which inherently encompasses essential details about road and building shapes, relative positions, and other attributes. Through the utilization of rendering techniques, the model seamlessly integrates external data and features such as symbol markers, drive trajectories, heatmaps, and user-defined markers, eliminating the need for extra feature engineering. The text module of GeoDecoder accepts various context texts and question prompts, generating text outputs in the style of GPT. Furthermore, the GPT-based model allows for the training and execution of multiple tasks within the same model in an end-to-end manner. To enhance map cognition and enable GeoDecoder to acquire knowledge about the distribution of geographic entities in Beijing, we devised eight fundamental geospatial tasks and conducted pretraining of the model using large-scale text-image samples. Subsequently, rapid fine-tuning was performed on three downstream tasks, resulting in significant performance improvements. The GeoDecoder model demonstrates a comprehensive understanding of map elements and their associated operations, enabling efficient and high-quality application of diverse geospatial tasks in different business scenarios.*

## 1. Introduction

Large language [5, 30, 34] and multimodal models [2, 18, 26] have gained considerable attention in academia and industry. However, their practical implementation in specific domains poses challenges. For example, in map and geographic tasks, using models like GPT3.5 to plan a trip to Beijing may result in unrealistic itineraries, such as visiting the Great Wall in the morning and the Forbidden City, located 80 kilometers away, in the afternoon. Lacking geographic context, such plans are unsuitable for direct user service. Therefore, leveraging domain-specific data and pretraining is essential to equip these models with the knowledge and capabilities to effectively understand and navigate geographic maps, bridging the gap between models and practical application in specialized domains.

The field of map and geographic information provides various services, including POI creation, recommendation, geocoding, navigation, and more [37, 39]. These services require different data types, features, and models to support their distinct outputs. For POI position generation, multiple channels gather information, including data from high-precision vehicles, user/business-provided information, and address inference from waybills. Factors considered in generating POI coordinates include the distribution of coordinates from different channels, data source reliability, and spatial attributes of the target area. Arrival point generation services employ text models to match POI names and addresses with the actual destination entity. Determining the drop-off location considers road and building block morphology, user trajectory, and dwell-time distribution. Implementing these services involves different teams using distinct datasets and models, resulting in significant costs for design and maintenance.

This paper introduces GeoDecoder, a dedicated multimodal pretrained model for the map and geographic information industry. GeoDecoder is trained on approximately 1.74 million POIs and 16,000 roads in Beijing, using a dataset of 22 million multimodal samples comprising text and images. Pretraining on eight map and geographic-related tasks enables GeoDecoder to develop a foundational understanding of the relative positions of landmarks, POIs, and roads in Beijing. GeoDecoder can align latitude and longitude, pixel coordinates, map markers, AOI morphology, surrounding geographic environment, roads, popular POI and AOI names and address. With this cognitive map as its foundation, GeoDecoder easily adapts to various downstream tasks, providing accurate and professional map services aligned with human geographic cognition.

The current trend in the industrial sector is to leverage a unified, large-scale model that can handle diverse types of data inputs and generate specific results to cater to different business requirements. GeoDecoder serves as a platform that aligns with this trend. Built on the beitGPT architecture [29, 35], GeoDecoder supports both image and text modalities as inputs for comprehensive data processing within the model. For instance, it can utilize base map to extract fundamental information about the shapes and relative positions of buildings and roads. Trajectories are represented with gradient-colored sequences of points to illustrate temporal sequences. Visual elements like symbols or logos indicate POI categories and user types on the map. Contextual information for tasks, such as POI names, addresses, and query prompts, are integrated through the text modality. GeoDecoder inherits the strengths of beit3 and uses separate expert modules for image and text modalities, enhancing the accuracy of intermodal interactions. By outputting results in GPT format, GeoDecoder can be deployed to provide unified model-based services for various tasks through prompt tuning.

## 2. Related work

Large language models (LLMs) like ChatGPT [25], LLAMA [42], ChatGLM [22, 41], and Qwen [1] are general-purpose NLP models trained on extensive text data from various sources. They excel in language comprehension, reasoning, and knowledge. However, describing map and geographic knowledge solely through language presents challenges. These models struggle with quantifying spatial concepts like distance and direction, and lack expertise in map understanding, limiting their proficiency in specialized geospatial services. While multimodal models like Kosmos [17, 21, 27], Blip [20, 40], and GPT4V [40, 23] may recognize common map elements like mountains, rivers, and labels, they lack specific training on map and geographic data, including roads, POIs, and relevant tasks, resulting in limited real-world geographic knowledge and cognition.

In recent years, Research institutions have explored large-scale models in the field of map and geographic information. For example, Ernie-Geo [16, 31, 32] combines heterogeneous graphs and text-based transformers to incorporate external POIs into the Ernie model. However, the limited range of data types introduced through this graph, primarily based on POI-POI co-location, query-click, and start-end POI, restricts its applicability to narrow business scopes like POI recommendations. MGeo [8] goes further by incorporating distance-based retrieval of relevant surrounding POIs and embedding their relative coordinates. Nevertheless, this model also faces limitations in handling diverse input data types, such as dynamic vehicle trajectory and pedestrian data, building and road morphology, and POI types and statuses. These structural limitations hinder the widespread promotion and development of both types of models in the map and geographic information field.

## 3. Method

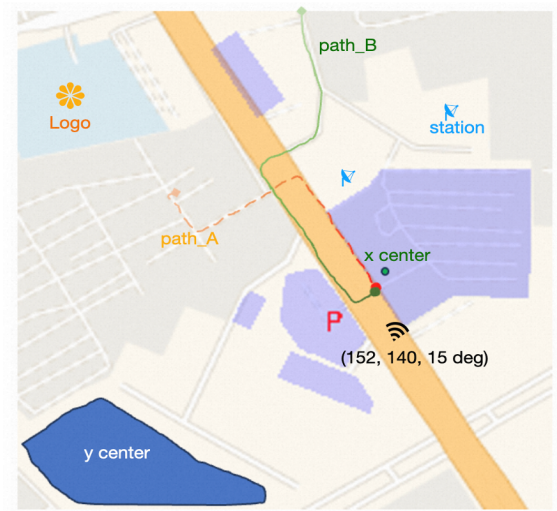

Figure 1: Visual input and visual pointing. The GeoDecoder supports the processing of various data types and their rendering onto the base map for comprehensive feature processing and semantic understanding. The base map provides valuable information regarding the structure and location of roads, buildings, green areas, rivers, and more. Trajectories and routes can be depicted using directed lines, while polygons can define larger entities or geographic boundaries. Smaller entities can be depicted using markers, such as dots, symbols, and logos. The GeoDecoder also allows for marker customization. For example, the black scanning marker in the image represents the pixel coordinates and shooting angle of a single image capture.

Based on our practical experience, we have observed that the input data for the majority of map and geographic-related businesses can be seamlessly integrated into the model through two primary channels: images and text. This integration allows for the direct rendering of a diverse array of static and dynamic data onto maps, which in turn enables the geographic model to process the data seamlessly, eliminating the need for laborious and time-consuming feature engineering and maintenance tasks. For instance, in Figure 1, driving and walking routes are represented using solid and dashed lines respectively, while the starting and ending points are denoted by diamond and circular markers. To distinguish between different user IDs, colors are employed, and the progressive shading of colors serves to indicate the temporal sequence of movement. Our observations further indicate that models are adept at acquiring a comprehensive understanding of the semantic information embedded within maps through pre-training and fine-tuning methodologies. Additionally, the inherent base map itself encompasses invaluable and rich geographic environmental information, encompassing details such as the morphology, relative positions, size, and orientation of roads, buildings, mountains, rivers, and more. In the past, the processing of such data necessitated specialized inputs in the form of data links and dedicated feature engineering. However, our GeoDecoder exhibits an exceptional capability for professional and simultaneous processing of multiple content types, akin to human cognition, while effectively considering their intricate topological relationships. Furthermore, the alignment of image-text modalities can be achieved seamlessly through the utilization of visual referencing techniques [40]. For example, the GeoDecoder accurately identifies that the term "blue AOI" corresponds to the region situated in the lower right corner of Figure 1, while the green dot represents the x center in the middle of the image. Symbol "P" denotes nearby parking lots, and light blue icons precisely delineate the locations of signal base stations. Our system also affords the flexibility to customize symbols to meet the specific requirements of individual businesses. Notably, the scanning marker depicted in Figure 1 signifies the position and shooting angle during high-precision vehicle photo captures, which play an indispensable role in calculating accurate coordinates for POI production. Therefore, our GeoDecoder effectively facilitates the rendering of diverse data types onto maps, enabling large-scale models to gain a profound comprehension of their underlying meanings and associated operations through end-to-end training. In doing so, we are able to streamline the laborious processes associated with feature engineering and model design, among other complex tasks.

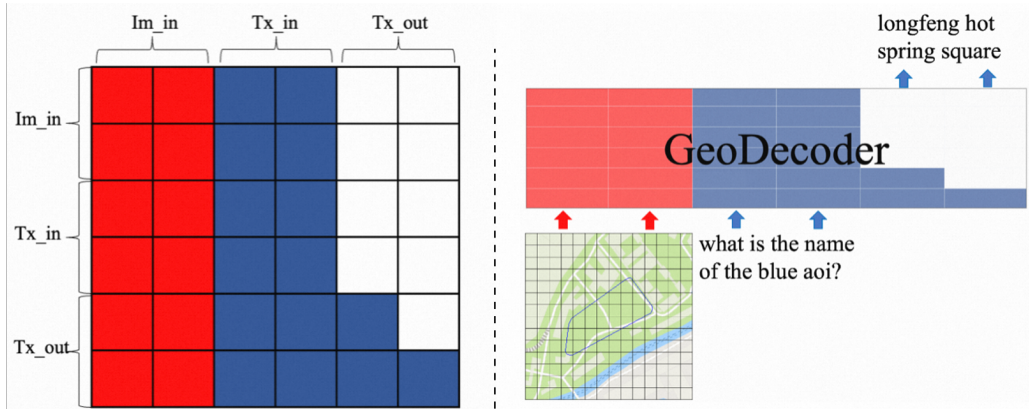

Figure 2: GeoDecoder Architecture. The GeoDecoder model, based on beitGPT, comprises three main modules: the image expert processing module (red), the input text processing module (blue), and the GPT-style text output module (blue-white). The model partitions the map, which includes various rendered information like AOI contours, into 14*14 patches. The input text and image exhibit bidirectional attention, similar to BERT, facilitating mutual comprehension. Word-by-word generation of the output text is achieved using attention masks.

Text is another important input modality that is more suitable for certain types of information, such as POI names and addresses, to be processed by the model in textual form rather than as images. With the professional text processing capability of GeoDecoder, we can perform tasks such as POI categorization, analysis of parent-child relationships, and address tree segmentation and entity recognition on these text inputs. Additionally, contextual information related to the task, spatial textual relationship descriptions, and problem statements often need to be provided through text input as well. For example, questions like "Where is the convenient parking spot if you work at y_center?" or "Does the green trajectory reach the north gate of x_center?" require text-based input to convey the relevant context and spatial relationships.

As shown in Figure 2, the GeoDecoder model is based on the beitGPT architecture and is specialized in separate processing for image input (Im_in), text input (Tx_in), and text output (Tx_out). Considering that the image module needs to first identify color, texture, extract boundary shapes, and then recognize objects and states [10, 14, 33], while the text module needs to map token IDs to embeddings, understand the meaning of associated phrases, analyze grammar, and then comprehend sentence meaning [6, 19, 22], there are significant differences in the processing modes of the two modalities. It is clearly inefficient and unreasonable to directly use the same QKV and FFN parameters to process hidden vectors of the two modalities [36]. Moreover, the human brain also processes vision and audition separately, using different regions such as the primary visual cortex (V1), secondary visual cortex (V2, V3), and temporal cortex for visual features like shape, color, and motion recognition[4, 7, 13], while the primary auditory cortex (A1) and secondary auditory cortex (A2) are involved in language processing [3, 15], with each region having different functions for distinguishing auditory features such as frequency, intensity, and direction. Wernicke's Area [12] is responsible for syntactic parsing and language comprehension, while Broca's Area [28] is responsible for language output. This differentiated processing of multiple modalities has also shown significant improvements in experimental results compared to unified processing [35].

The input side of the method incorporates a bidirectional attention mechanism similar to BERT, which plays a crucial role in facilitating mutual understanding and interaction between text and image [41]. By allowing each token input to access both preceding and succeeding information, the former image embedding can take into account the focus of the latter text. Moreover, image processing should be a globally unified and simultaneous process, regardless of the direction of processing. Whether starting from the top left corner to the bottom right corner or vice versa, the output should remain consistent regardless of the model's processing order. In contrast, the text output side operates differently. Language, unlike images, follows a temporal sequence where the preceding output content influences the subsequent content. To achieve GPT-style word-by-word output, a lower triangular attention mask is employed [29]. This approach offers significant advantages, including the absence of length and format constraints in the output. It enables the model to generate results in a free-text format, tailored to the requirements of any given task. By establishing the correlation between image and text and employing result inference through next word prediction, as opposed to the blank filling approach used in BERT, the method effectively and flexibly supports various types of geospatial image-text tasks.

Mathematically, we can articulate the entire process as follows. Given $X_{in} = concat(X_{im}, X_{tx}) \in R^{bs*(D_{im}+D_{tx})*d}$ the attention layer's input hidden states situated within the framework defined by batch size ($bs$), image, and text lengths ($D_{im}$ and $D_{tx}$), as well as the dimension of the hidden state ($d$), the procedure begins with $X_{in}$ undergoing layer normalization. This step prefaces the self-attention phase as delineated in equations (1-2):

$$X' = MSA(LN(X_{in})) + X_{in} \quad (1)$$

$$MSA(X_{in}, W_{im}, W_{tx}) = softmax(\frac{mask(QK^\top)}{\sqrt{D}})V \quad (2)$$

The QKV matrices are assembled as follows:
$$Q = concat(X_{im} \cdot W^Q_{im}, X_{tx} \cdot W^Q_{tx})$$
$$K = concat(X_{im} \cdot W^K_{im}, X_{tx} \cdot W^K_{tx}) \quad (3)$$
$$V = concat(X_{im} \cdot W^V_{im}, X_{tx} \cdot W^V_{tx})$$

Here, $mask(\cdot)$ refers to the masking function derived from the GeoDecoder arrangement in figure2a. Post normalizing again to transition from X' to X", the FFN layer merges visual and textual representations to produce the final output informed by

$$X_{out} = concat(FFN_{im}(X"_{im}), FFN_{tx}(X"_{tx})) + X" \quad (4)$$

The overall data processing pipeline follows these steps. Firstly, various types of data required for the task are either rendered onto a map or inputted to the model via the text side, based on specific business requirements. The image is prepared to 224px*224px and divided into 14*14 patches, with each patch measuring 16px*16px. These patches undergo convolutional operation to generate image embeddings of dimension 196*768. On the other hand, the text data is tokenized into IDs using a vocabulary size of 82088, primarily consisting of Chinese characters and English words. These IDs are then mapped to their corresponding text embeddings. These text embeddings, along with the image embeddings, are added with position embeddings and fed into the GeoDecoder model for further processing. Considering computational costs and the need for high throughput in daily business operations, we conducted experiments using a default configuration of a 12-layer beitGPT model, which consists of 297M parameters. Within each layer, the multihead self-attention mechanism's QKV, output projection, layer normalization, and feed forward operations are processed separately for each modality. At the text output side, a triangular attention mask is added to achieve unidirectional attention. After passing through the 12 layers of Transformers, we adopt a word-by-word approach to generate text outputs. This is accomplished by applying softmax to obtain the word with the highest probability controlled by a temperature parameter. The generation process continues until the <\s> (end of sentence) token is encountered.

## 4. Pretraining tasks

The geospatial multimodal pretraining tasks aim to facilitate the model's acquisition of geographic knowledge and enhance its understanding of geographical concepts. These tasks encompass a range of aspects, including the recognition of geographic elements on maps, comprehension of the common meanings associated with different symbols, interpretation of entity references in text overlaid on images, and identification of entity names within images. Additionally, these tasks involve aligning marked points on the map with their corresponding latitude, longitude, and address information. To further enhance the model's capabilities, the pretraining tasks incorporate commonsense knowledge, such as the expectation that car trajectories should follow roads, while shops or hotels should not be located on roads or in water bodies. Moreover, the tasks include the acquisition of geographic knowledge about specific regions of interest, such as the northwestern corner of Beijing where Peking University is situated. To accomplish these objectives, we have carefully curated eight distinct task types, as illustrated in Figure 3, comprising approximately 22 million image-text pairs sourced from 1.7 million POIs and 16,000 road segments in Beijing. The ultimate goal is to equip the GeoDecoder

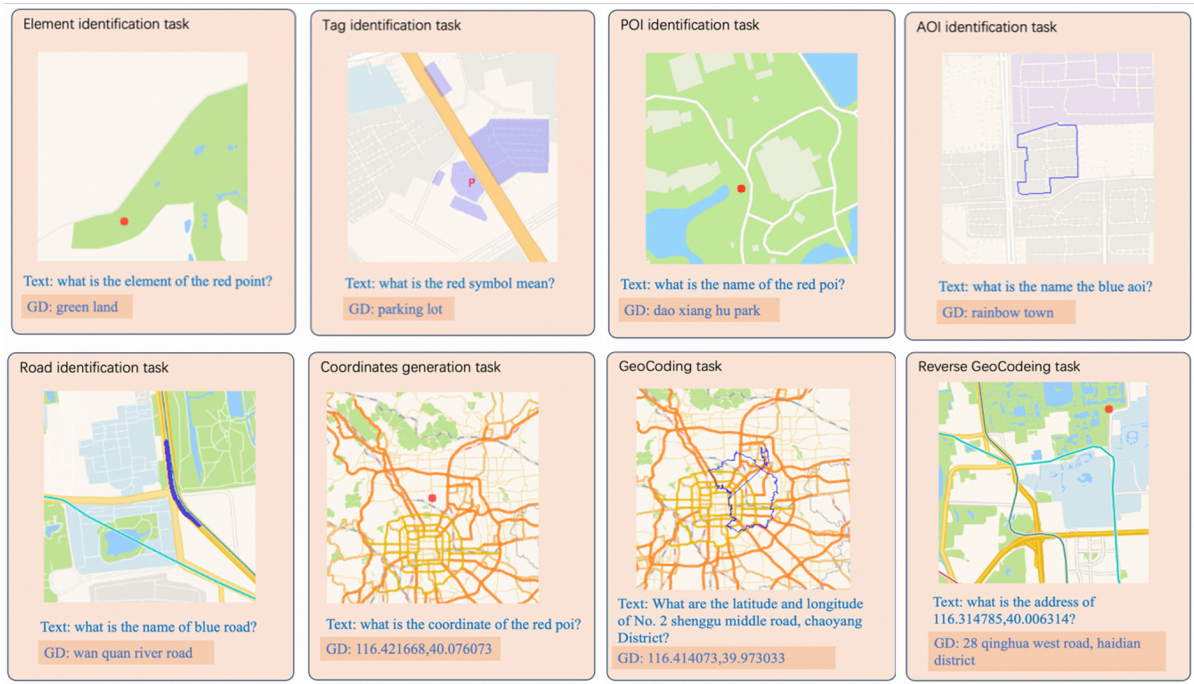

Figure 3: Eight GeoDecoder (GD) pre-training tasks. Element identification task, GD needs to identify the geographic element at the marked locations. Tag identification task, GD needs to recognize the meanings of symbols or logos indicated by the text input. POI identification task, GD needs to determine the names of entities based on the marked locations and the surrounding geographic context. AOI identification task, GD needs to determine the names of entities based on the AOI shape and the surrounding geographic context. Road identification task, GD needs to provide the names of roads based on the highlighted road segments and the surrounding geographic context. Coordinates Generation task, GD needs to generate the corresponding latitude and longitude based on the marked locations and the surrounding geographic context. Geocoding task: GD needs to provide the corresponding latitude and longitude information based on the text address and POI names. Reverse geocoding task: GD needs to provide the corresponding text address based on the latitude and longitude and the surrounding geographic context of the map marked locations.

| Tasks | Element identification | Tag identification | Poi identification | Aoi identification | Road identification | Coordinate generation | geocoding | Reverse geocoding |
|---|---|---|---|---|---|---|---|---|
| **Samples** | 296,636 | 495,309 | 5,725,200 | 4,618,350 | 668,622 | 1,324,625 | 5,198,512 | 3,858,043 |
| **Accuracy** | 98.3% | 99.9% | 57.6% | 58.6% | 85.9% | 305m | 107m | 78.3% |

Table 1: 8 pretraining tasks along with their training sample size and accuracy. The accuracy of the coordinates generation task and geocoding task is indicated by the median prediction error.

with essential geographic knowledge and enhance its geographic cognition through pretraining.

For the Element identification task, the GeoDecoder is tasked with identifying the geographic element at the marked location based on the image. The model needs to acquire an understanding of the color and dot mentioned in the text and locate the corresponding entity within the image. Additionally, it must determine the geographic feature at the marked point by considering the surrounding geographic context. There is a diverse range of geographic elements, which can exhibit variations in shape and color depending on their classification, grade and map scale. For example, major roads and minor roads may be represented differently on the map. We have prepared a dataset of 297,636 image-text pairs specifically for this task, with an overall accuracy of 98.3% in the element identification task.

For the Tag identification task, the GeoDecoder's objective is to identify symbols within the image based on their colors, recognize the shape of these symbols, and determine their corresponding meanings. The model is then expected to generate textual descriptions of the symbols as output. For instance, the symbol "P" may indicate a parking lot, while "G" might represent a gate. In addition to supporting simple letters and logos, this task also encompasses user-defined markers and the associated meanings. An example of such a marker is the scanning marker depicted in Figure 1, which can convey information about the location and angle of image capture. To train the model in comprehending fundamental map-related symbols, we have meticulously curated a dataset of 495,309 image-text pairs tailored specifically for this task. Remarkably, the GeoDecoder achieves an impressive accuracy of 99.0% in the Tag identification task.

For the poi identification task, the GeoDecoder's objective is to recall and determine the name of the poi based on the marked point and surrounding geographic context. Considering the high density of pois in many areas, we structured the frequency of poi appearances in the training dataset according to their popularity in poi search records. This approach ensures that the model acquires a memory and recognition of prominent landmarks while potentially overlooking less common pois. We have curated a dataset of 5.7 million image-text pairs specifically designed for this task. However, the accuracy of poi identification only reaches 57.6%. The majority of errors occur when the model confuses the names of pois with highly popular pois located in close proximity. Nevertheless, overall, we can still observe the GeoDecoder's ability to recall and make reasonable judgments based on the geographic environment of the map. For detailed examples of errors and strategies for improvement, please refer to the appendix.

For the aoi identification task, the GeoDecoder's objective is to determine and output the name of the aoi based on the shape of the selected aoi and the surrounding geographic context. We have prepared a dataset of 4.6 million image-text pairs specifically tailored for this task. The accuracy of aoi identification is 58.6%, slightly higher than poi identification. This improvement can be attributed to the inclusion of aoi shape information, despite having 19% fewer training samples compared to poi identification. The main errors in the output occur when the model confuses the names of highly popular aois in the vicinity and when the map scale of certain aois is too small, resulting in insufficient surrounding geographic information to determine their names.

For the road identification task, the GeoDecoder's objective is to determine and output the corresponding road name based on the highlighted road segment in the image and the surrounding geographic context. We have curated a training dataset of 668,622 image-text pairs specifically designed for this task. The accuracy of road identification reaches an impressive 85.9%, which is significantly higher than the accuracy of poi and aoi identification. This improvement can be attributed to the relatively smaller number of road entities compared to pois and aois, making it easier for the model to remember their distinctive visual features. Additionally, the surrounding geographic environment of roads often provides clear indicators, resulting in relatively straightforward identification of road segment names. As a result, the vast majority of road name outputs closely align with the correct answers, highlighting the GeoDecoder's highly accurate memory and cognitive abilities when it comes to roads.

For the coordinates generation task, the geocoder aims to determine and provide the latitude and longitude based on the marked point location in the image and the surrounding geographic context. We have prepared a dataset of 1.3 million image-text pairs specifically designed for this task. The final median prediction error for this task is 305 meters. The main source of this error is attributed to the use of a constant image scale of 11, where the pixel distance is approximately 100 meters, resulting in an error of around 3 pixels for the median prediction error. Additionally, the prediction error is significantly influenced by the prediction of high-order digits. For example, in one sample, the true coordinates are 116.519630, 39.774726, while the predicted coordinates are 116.52926, 339.7740, indicating an error in the latitude prediction that leads to a significant distance error, despite the relatively accurate prediction of other digits. Errors exceeding 1 kilometer account for 12.8% of cases, primarily caused by incorrect predictions of high-order digits. Detailed strategies for improving prediction accuracy and methods for error detection can be found in the appendix.

For the geocoding task, the geocoder's objective is to recall and determine the corresponding latitude and longitude coordinates based on the provided poi address and name in the text. Additionally, we have provided the county-level aoi geographic boundaries based on the poi addresses to help reduce the occurrence of large-scale prediction errors. We have prepared a training dataset of 5.2 million image-text pairs specifically for this task. The median prediction error for geocoding is 107 meters, with 48.5% of the predictions having errors within 100 meters. The higher accuracy compared to the coordinates generation task is primarily due to the significant increase in training sample size and the mutual reinforcement between geocoding and reverse geocoding tasks. Among the cases with larger errors in latitude and longitude predictions, they account for 12.8%, primarily caused by errors in predicting high-order digits.

For the reverse geocoding task, the GeoDecoder aims to determine and predict the corresponding textual address based on the marked point in the image, the surrounding geographic context, and the latitude and longitude information in the text. We have prepared a training dataset of 3.9 million image-text pairs specifically tailored for this task. The accuracy of address prediction reaches 78.3%. The main errors are concentrated in cases where the model predicts incomplete address information, such as only providing the city or county without including street names or building numbers. Based on the test results, we have observed that the GeoDecoder demonstrates relative accuracy in associating latitude and longitude with corresponding location address information.

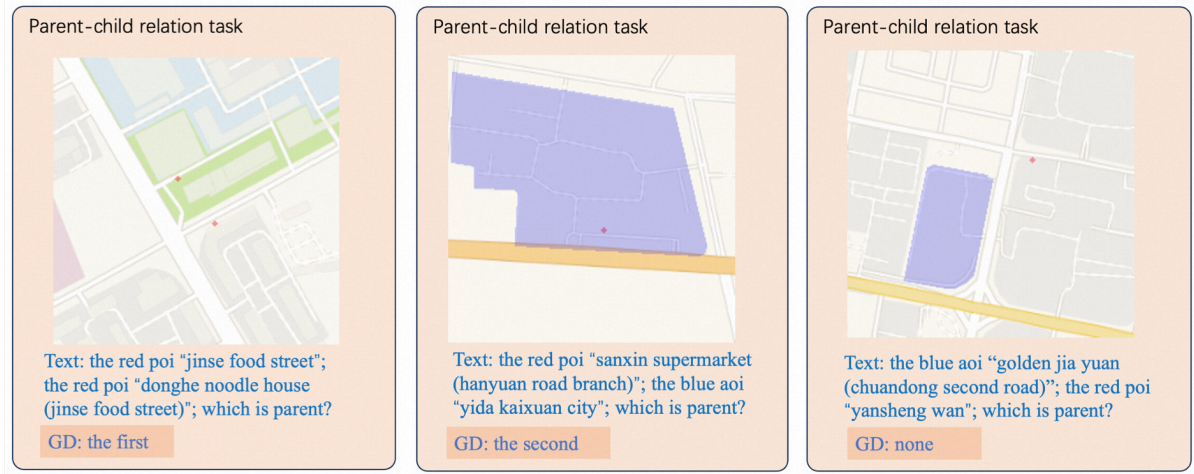

Figure 4: Parent-child relation task. The image displays two POI or AOI, while the text provides the names of the corresponding entities. The GeoDecoder's task is to analyze the image-text information and determine if there exists a parent-child relationship between the entities. Furthermore, if such a relationship is present, the GeoDecoder must identify which entity serves as the parent.

## 5. Finetuning tasks

After 2 weeks of pretraining on 22 million image-text pairs, the GeoDecoder has developed a solid understanding of map cognition and the geographical layout of Beijing. Leveraging this foundation, we proceeded with rapid finetuning on three downstream business tasks, each of which produced results that surpassed our initial expectations.

Parent-child relationship judgment plays a crucial role in various map-related businesses, including the construction of a POI knowledge graph and even the creation of POIs themselves. In many cases, data from different sources may point to the same POI entity, requiring us to match and classify them based on their names and distances. The primary cause of POI creation mistakes is parent-child mismatch, such as the Department of Electronics at Peking University and the Quantum Electronics Laboratory at Peking University. These entities have similar names and are in close proximity, often resulting in frequent mismatches by the model. Consequently, the information of both entities becomes mixed and displayed as a single POI on the map, leading to severe consequences. To address this issue, we specifically optimized the parent-child relationship task using GeoDecoder finetuning. In this task, we provided the model with the visual spatial relationship and shape information of the AOI or POI for the two entities, as well as the textual content of their names, as shown in Figure 4. The model was trained to utilize the topological relationship of the two entities in the image and the implied relationship of their textual names to accurately determine the parent-child relationship and identify the parent entity during finetuning. We prepared a high-quality dataset of 5M image-text pairs representing parent-child relationships for finetuning. The training process took 12 hours, utilizing both from scratch GeoDecoder and pretrained GeoDecoder models trained on four A6000 GPUs for three epochs. To evaluate the models, we used a test set of 100,000 unseen parent-child relationships, with the experimental results presented in Table 2. The accuracy for the three scenarios (no parent-child relationship, parent in front, and parent at the end) without pretraining were 89.1%, 91.0%, and 88.7%, respectively. With pretraining, the accuracy significantly improved to 96.5%, 98.2%, and 99.1%. The experimental results demonstrate that (1) the pretrained model's map geography knowledge and cognitive abilities significantly enhance the performance of the parent-child task, (2) GeoDecoder exhibits remarkable multimodal processing capabilities for images and text, and (3) downstream businesses can efficiently and effectively implement GeoDecoder-based solutions with ease and high quality.

In the POI coordinate generation task, we render coordinate data from different information channels onto the map, with colors representing the information sources, and the corresponding POI names provided in the text, as illustrated in Figure 5. GeoDecoder learns to utilize the map environment information and ground truth guidance to judge and generate the most reasonable coordinate positions. This requires the model to determine which information sources and channels provide more reliable coordinates. For example, coordinates collected by cameras attached to vehicles are often closer to the surface of buildings, while coordinates obtained through geocoding from shipping documents may lack precision and both are not suitable as display coordinates. On the other hand, for buildings and scenic POIs, the geometric center is often used as the coordinate position, and coordinates for POIs inside buildings are close to the corresponding Wi-Fi locations. We prepared 5.60M training samples and finetuned the model for three epochs over half a day for the POI coordinate generation task. When comparing with the online rule-based coordinate generation, we observed that GeoDecoder achieved a proportion of 78.0% for cases with a distance error range within 20 meters, while the online version only accounted for 65.4%. For cases where the error distance exceeded 100 meters, GeoDecoder accounted for only 9.8%, whereas the online version accounted for 17.1%. This finetuning task in the coordinate generation demonstrates GeoDecoder's ability to understand task requirements and textual content, utilize geographical information from the map, and summarize experience from color-coded source defined by us to make accurate business judgments.

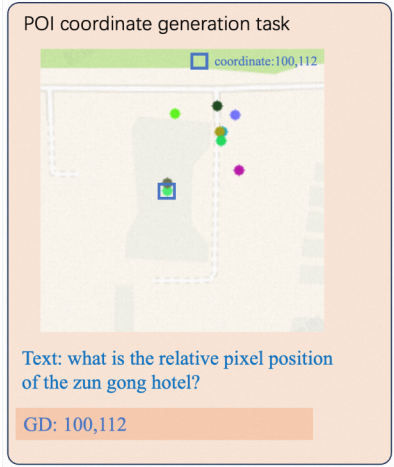

Figure 5: POI coordinate generation task, different information sources (represented by different colors) provide different coordinates marked on the map. GD generates the most reliable POI display coordinates based on the provided image-text information and its expertise. The blue box here is rendered based on GD's textual output.

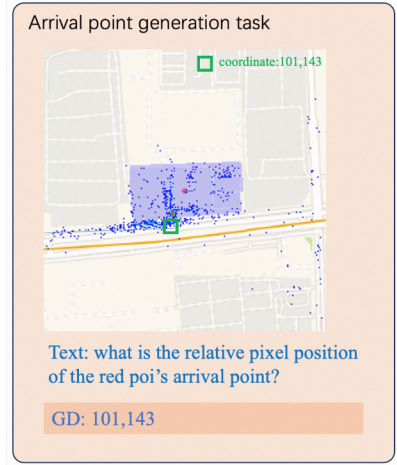

Figure 6: Arrival point generation task, POIs and associated user dwellings (represented by a heatmap) are presented on the map. GD generates the most suitable arrival point coordinates based on the provided image information and its experience. The green box here is rendered based on GD's textual output.

In the arrival point generation task, we visualize the POI markers and user dwellings as a heatmap on the map and input the corresponding POI names and queries as text into Geocoder, as shown in Figure 6. Geocoder analyzes the provided information and ground truth data to accurately determine and output the pixel coordinates of the arrival points. To achieve this, we curated a dataset of 2 million image-text pairs for training and conducted finetuning on GeoDecoder for 18 epochs in a single day. Performance evaluation was done on 180,000 previously unseen test samples. The online road-based algorithm achieved an accuracy of 73.0% (measured by arrival index in the appendix), while the Segformer [38] model achieved an accuracy of 73.2%. With finetuning based on pretrained GeoDecoder, the accuracy significantly improved to 76.7%. Furthermore, incorporating additional user dwelling information in the form of a heatmap increased the accuracy to 79.8%. These results demonstrate that GeoDecoder quickly adapts to the arrival point generation task through finetuning and effectively utilizes heatmaps, which were not included in the pretraining, to enhance the task. This approach enables the incorporation of various forms of external information into the model, allowing it to quickly learn and leverage the information through finetuning, without the need for specific feature engineering for each type of information.

# 6. Conclusion

GeoDecoder, based on the BeitGPT architecture, is introduced for geospatial tasks. The model incorporates separate expert modules to process images and texts, and employs a GPT-based structure to generate diverse task outputs. The underlying base map contains comprehensive information about building and road structures, as well as their relative spatial relationships. The model enables the integration of customized data and features onto the map to enhance task execution. On the textual side, various inputs including POI names, addresses, task context, and queries are utilized. To enhance the model's understanding of fundamental map elements, relationships, and the layout of geospatial information in Beijing, we employed eight different pretraining tasks for GeoDecoder. Furthermore, fine-tuning was performed on tasks such as parent-child relation discrimination, POI coordinate generation, and POI arrival point generation, resulting in outcomes that surpassed our initial expectations. Overall, GeoDecoder proves to be a strong base model for map-related tasks.

# Appendix

## 7. POI identification task

Figure 7 illustrates a typical bad case in the POI identification task. Based on the geo-tagged image and surrounding geographical information, GeoDecoder incorrectly predicted the location of "Weisheng Mansion" as "QiDi Technologies". When we searched for "QiDi Technologies", we found that it is actually located 20 meters to the right of Weisheng Mansion, as shown in Figure 7. One significant reason for this erroneous prediction is that the frequency of occurrence in the training samples is higher for "QiDi Technologies" (ranked 104,397th in national popularity) than for "Weisheng Mansion" (ranked 443,991st) due to its higher search popularity. In future versions, to improve the quality of POI identification, we will (1) employ larger model parameters to enhance the model's memory and decision-making abilities, (2) use more training samples and longer training steps to achieve convergence at lower loss levels, and (3) incorporate contrastive learning to enhance the model's entity distinguishability.

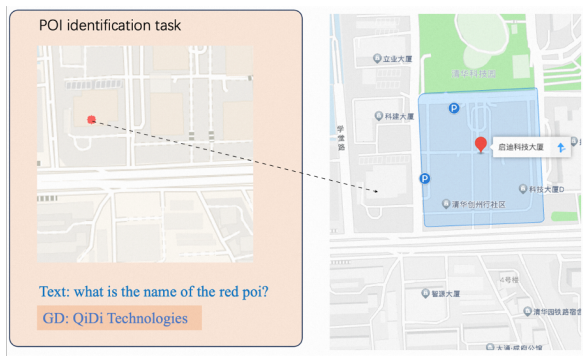

Figure 7: An unsuccessful case in the POI identification task. GeoDecoder predicted "QiDi Technologies" while the actual Ground Truth name is "Weisheng Mansion". However, the light blue AOI displayed on the right side of the figure represents "Weisheng Mansion".

## 8. Coordinates generation task

The task of generating coordinates presents significant challenges mainly because the pretraining task only supplied a map with a scale of 11, leading to an approximate distance of 100 meters between adjacent pixels. Table 2 illustrates the distribution of distance errors observed in the final test data. Successfully completing this task necessitates the utilization of large-scale maps (scale 9-11) to discern the high-order latitude and longitude digits at the provinces and cities level, as well as small-scale maps (scale 12-18) to pinpoint the specific location of markers, representing the low-order latitude and longitude digits. In the future, we plan to incorporate techniques such as Perceiver, allowing the model to leverage multiple feature maps for richer input and more accurate prediction outputs.

| Distance level (m) | Percentage |
|---|---|
| 0 ~ 100 | 27.7% |
| 100 ~ 200 | 11.7% |
| 200 ~ 500 | 21.3% |
| 500 ~ 1000 | 26.6% |
| 1000 ~ | 12.8% |

Table 2: Distribution of error distance in the coordinates generation task.

## 9. Geodecoding task

| Distance level (m) | Percentage |
|---|---|
| 0 ~ 100 | 48.5% |
| 100 ~ 200 | 11.0% |
| 200 ~ 500 | 14.5% |
| 500 ~ 1000 | 10.5% |
| 1000 ~ | 15.5% |

Table 3: Distribution of error distance in the geodecoding task.

## 10. Parent-child relation task

| label | GD finetune (wo/ pretrained) | GD finetune (pretrained) |
|---|---|---|
| No parent-child relation | 89.1% | 96.5% |
| The first one is the parent | 91.0% | 98.2% |
| The second one is the parent | 88.7% | 99.1% |

Table 4: Results of the parent-child relation task. The experiment compared the performance of directly finetuning the task from scratch and finetuning the task from a pretrained model. The latter approach showed significant improvements in all three scenarios.

## 11. POI coordinate generation task

| Distance | Online | GeoDecoder |
|---|---|---|
| = 0 | 25.1% | 46.1% |
| 0 ~ 20 | 40.3% | 31.9% |
| 20 ~ 50 | 11.7% | 7.8% |
| 50 ~ 100 | 5.8% | 4.4% |
| 100 ~ | 17.1% | 9.8% |

Table 5: Distribution of error distances in the POI coordinate generation task. The middle column illustrates the error distribution of predictions based on the online service utilizing heuristic rules, while the third column displays the error distribution of predictions based on the GeoDecoder model after undergoing fine-tuning.

## 12. Arrival point generation task

| Strategies | Arrival Index |
|---|---|
| Road-based algorithm (online) | 73.0% |
| FCN | 62.7% |
| SegFormer | 73.2% |
| GeoDecoder finetune | 76.7% |
| GeoDecoder + heatmap finetune | 79.8% |

Table 6: Performance comparison of different strategies in the task of generating arrival points. The strategies evaluated include the road-based algorithm used in the online service, FCN (Fully Convolutional Network), SegFormer, GeoDecoder with Finetuning, and GeoDecoder with Finetuning combined with user-dwellings heatmaps.

Arrival point index is defined as follows:
• 1 point: Predicted value falls on the same road as the ground truth, with a distance less than 30 meters.
• 0.5 points: Predicted value falls on the same road as the ground truth, with a distance less than 50 meters.
• 0 points: Other cases.
This arrival index reflects the quality of arrival point generation.

## 13. Model hyperparameters

| Model | GeoDecoder |
|---|---|
| Training Dataset | 22 Million samples: 1.74 Million POIs 16,000 Roads in Beijing |
| Training Parameters | Batchsize: 64 |
| | Epochs: 20 |
| | Trained for: 2 weeks |
| | GPU: 4 x A6000 |
| Image Input | 224px x 224px |
| | 14 patches x 14 patches |
| | Patch size: 16px x 16px |
| Token Count | 196 for image |
| | 60 for text |
| | Vocabulary Size: 82,088 |
| Parameters | 297 Million |
| Layers | 12 |
| Hidden Size | 768 |
| FFN Intermediate Size | 3072 |
| Attention Heads | 16 |
| Learning Parameters | Optimizer: AdamW |
| | Beta1: 0.9 |
| | Beta2: 0.98 |
| | Warmup Steps: 100 |
| | Peak Learning Rate: 1e-4 |
| | Learning Rate Schedule: Linear |
| | Weight Decay: 0.01 |
| | Dropout Rate: 0.1 |

Table 7: GeoDecoder model hyperparameters.